\documentclass[lettersize,journal]{IEEEtran}
\usepackage{amsmath,amsfonts}
\usepackage{algorithmic}
\usepackage{algorithm}
\usepackage{array}
\usepackage[caption=false,font=normalsize,labelfont=sf,textfont=sf]{subfig}
\usepackage{textcomp}
\usepackage{stfloats}
\usepackage{url}
\usepackage{verbatim}
\usepackage{graphicx}
\usepackage{algorithm}
\usepackage{enumitem}
\usepackage{multirow}
\usepackage{booktabs}
\usepackage{color}
\usepackage{hyperref}
\usepackage{epsfig}
\usepackage{orcidlink}
\hypersetup{hidelinks} 

\usepackage{amssymb}
\usepackage{lipsum}
\usepackage{epstopdf}
\usepackage{enumitem}
\usepackage{cite}
\begin{document}

\title{On Support Relations Inference and Scene Hierarchy Graph Construction from Point Cloud in Clustered Environments}

\author{Gang Ma$^{\orcidlink{0000-0003-3365-5188}}$ and Hui Wei$^{\orcidlink{0000-0003-2696-0707}}$
\thanks{The authors are with the Laboratory of Algorithms for Cognitive Models, Shanghai Key Laboratory of Data Science, School of Computer Science, Fudan University, Shanghai 200433, China (e-mail: mag20@fudan.edu.cn; weihui@fudan.edu.cn). \textit{(Corresponding author: Hui Wei.)}}}



\maketitle

\begin{abstract}
Over the years, scene understanding has attracted a growing interest in computer vision, providing the semantic and physical scene information necessary for robots to complete some particular tasks autonomously. In 3D scenes, rich spatial geometric and topological information are often ignored by RGB-based approaches for scene understanding. In this study, we develop a bottom-up approach for scene understanding that infers support relations between objects from a point cloud. Our approach utilizes the spatial topology information of the plane pairs in the scene, consisting of three major steps. 1) Detection of pairwise spatial configuration: dividing primitive pairs into local support connection and local inner connection; 2) primitive classification: a combinatorial optimization method applied to classify primitives; and 3) support relations inference and hierarchy graph construction: bottom-up support relations inference and scene hierarchy graph construction containing primitive level and object level. Through experiments, we demonstrate that the algorithm achieves excellent performance in primitive classification and support relations inference. Additionally, we show that the scene hierarchy graph contains rich geometric and topological information of objects, and it possesses great scalability for scene understanding. 
\end{abstract}



\section{Introduction}

\IEEEPARstart{S}{cene} understanding is one of the most popular and challenging research topics related to the development of depth-sensing devices. Traditionally, scene understanding refers to labeling each object in an image \cite{he2016deep}\cite{zhen2020joint}\cite{minaee2021image}, a task in which current methods have achieved excellent results. Scene understanding can also involve more high-level understanding and inference tasks, such as layout estimation \cite{liu2015rent3d}\cite{hayat2016spatial}\cite{yan20203d}, physical relations inference \cite{wald2020learning}\cite{kim20193}, and stability inference \cite{mojtahedzadeh2015support}\cite{zheng2013beyond}. Support relations inference has a high impact in many robot applications, particularly those wherein robots not only need to know the label information of objects in the scene but also the geometric information and the topological relationships between objects.

The scene graph is a powerful tool for higher-level scene understanding and inference \cite{chang2021comprehensive}, which can clearly express the objects, attributes, and relations between objects in the scene. Several studies \cite{johnson2015image}\cite{johnson2018image}\cite{schroeder2020structured} directly used it as the input for scene understanding. Yang et al. \cite{yang2017support} constructed a semantic scene graph from support relations inference in the scene and introduced some approaches to evaluate the quality of the generated semantic graph. Although the graph contains the position and support relations of objects in the scene, it is still crucial to exploit the richer geometric and topological information. For example, if a robot takes a cup from a cabinet, besides support information, the configurations of the cabinet and cup are also needed to calculate the optimal grasping pose.

A large number of basic shapes, such as cylinders, spheres, and especially planes, are usually contained in 3D scenes. However, the majority of the existing support relations inference studies based on RGB images ignore much of the spatial geometric and topological information in the scenes. The goal of our work is to make full use of information from the point cloud to infer the support relations in the scene from the bottom up and generate a scene hierarchy graph to describe the geometric information and the topological relationship between objects in the scene in a more fine-grained way. 

\begin{figure*}[tbp]
\centering
\includegraphics[width=\linewidth]{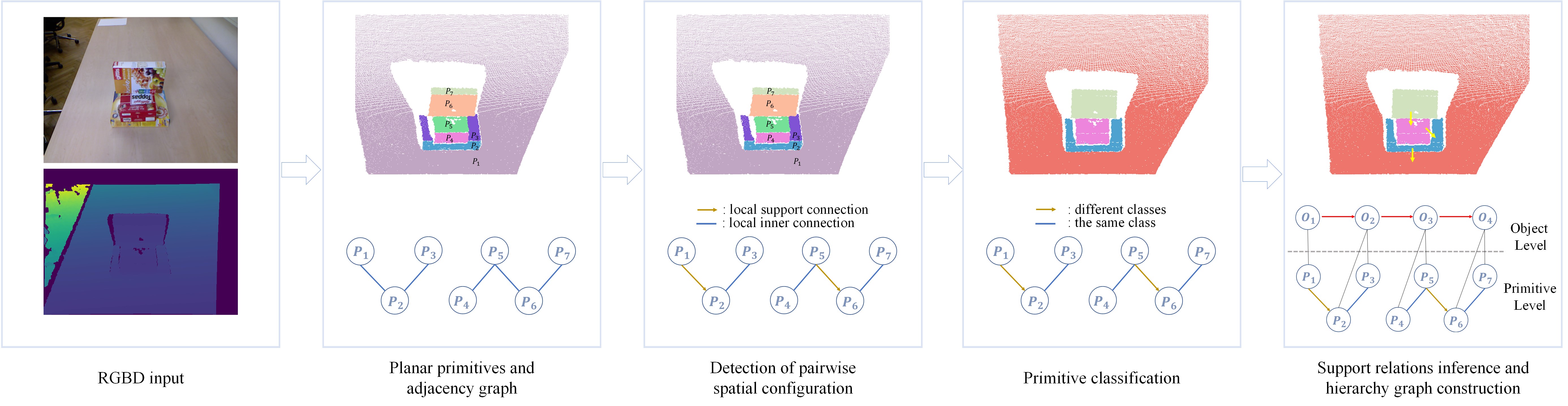}
\caption{%
An overview  of our approach. }
\label{fig:teaser}
\end{figure*}

 Our pipeline considers as input data a set of RGBD images. First, plane primitives are extracted from the scene and an adjacency graph is constructed. Next, the spatial configuration of the neighboring plane pair in the adjacency graph is detected and divided into local support connection and local inner connection. Then, a combinatorial optimization approach is used to classify the planar primitives and regard each label as an object. Finally, a hierarchical graph is constructed and support relations are inferred based on the segmentation results (see also Figure~\ref{fig:teaser}).
\IEEEpubidadjcol
To summarize, our contributions are:
\begin{itemize} 
\item[1.] An approach of spatial configuration detection based on plane pairs is proposed. With this approach, primitive pairs are first divided into two types: local support connection and local inner connection, and then combinatorial optimization is used to classify the planar primitives.

\item[2.] A bottom-up approach of support relations inference is proposed. With this approach, local support relations inference and global support relations inference are performed to more comprehensively utilize the spatial topology information of planar primitive pairs.

\item[3.] The use of a hierarchical graph is proposed to augment and finalize the support relations inference. With this approach, the scene hierarchy graph is constructed because it contains not only the rich geometric information of objects but also the spatial topological information between objects. It also shows good scalability.
\end{itemize}
This paper is structured as follows. Related work is discussed in Section~\ref{sec:related_work}. In Section~\ref{sec:detection_of_pairwise_spatial_configuration}, detection of pairwise spatial configuration is explained. A method to adopt a combinatorial optimization approach to classify the plane primitives is proposed in Section~\ref{sec:primitive_classification}. In Section~\ref{sec:support_relations_inference}, the inference of a scene hierarchy graph is explained. Experimental results of the proposed framework are shown in Section~\ref{sec:experiments}. Finally, a conclusion in Section~\ref{sec:conclusion}, summarizes this paper. 


\section{Related Work}
\label{sec:related_work}

Scene understanding has always been an important research topic in the development of computer vision. Some related studies on support relations inference and scene graph are introduced in this section.

\subsection{Support Relations Inference}
\label{subsec:sri_rw}

Silberman et al. \cite{ silberman2012indoor} proposed an approach to infer the support relations between objects of an indoor scene from an RGBD image. To better understand how 3D cues help to interpret a 3D structure, they provided a new integer programming formulation to infer physical support relations. In addition, they demonstrated that 3D scene cues and support relations inference can optimize object segmentation. However, their approach ignored small objects and physical limitations. Xue et al. \cite{xue2015towards} found that the support relations and structural categories of indoor images are intrinsically related to physical stability. Thus, they constructed an improved energy function utilizing the stability. By minimizing this energy function, they further inferred support relations and structural classes from indoor RGBD images. Zhuo et al. \cite{zhuo2017indoor} argued that there is a strong connection among instance segmentation, semantic labeling, and support relations inference. They formulated their problem as that of jointly finding the regions in the hierarchy corresponding to instances and estimating their class labels and pairwise support relations by exploiting a hierarchical segmentation. However, these approaches, like most approaches based on pixel-wise segmentation, ignored a large amount of geometric and topological information in 3D scenes.

Jia et al. \cite{yan20203d} proposed a method based on 3D volume estimation because they hypothesized 3D stereo inference would be conducive for scene understanding. Using RGBD data to estimate 3D cuboids, they obtained spatial information about each object to infer their stability. However, their method is only suitable for scenes with a few objects.

\subsection{Scene Graph}
\label{subsec:scene_graph_rw}

Scene graph is a powerful tool for scene understanding \cite{chang2021comprehensive}, which has been applied to such areas as image understanding and reasoning \cite{aditya2018image}\cite{shi2019explainable}\cite{zhang2019large}, 3D scene understanding \cite{kim20193}\cite{armeni20193d}\cite{rosinol2021kimera}, and human--object/human--human interaction \cite{zhuang2018hcvrd}\cite{li2020visual}. Based on support relations inference, Yang et al. \cite{yang2017support} constructed a semantic graph to explain a given image and put forward some objective measures to evaluate the quality of the constructed graph. However, those scene graphs contain only partial information about the positions and support relations between objects and ignore richer geometric and topological information. Mojtahedzadeh et al. \cite{mojtahedzadeh2015support}, using the static equilibrium principle in classical mechanics, described a method for faster extraction of the support relations between pairs of objects in contact. They also generated support relation graphs to assist in solving the decision-making problem of stable robot grasping. However, their method was based on the prerequisite that the configuration of the point cloud is available. 

In the present study, we make full use of the spatial topological information of the planar primitive pairs in the scene and infer the support relations in the scene bottom-up. The scene hierarchy graph we construct contains rich object geometric and topological information and has excellent scalability. Hence, it can be applied to future research on the problem of stable grabbing decisions.

\section{Detection of Pairwise Spatial Configuration}
\label{sec:detection_of_pairwise_spatial_configuration}
\begin{figure}[htb]
\centering
\includegraphics[width=\linewidth]{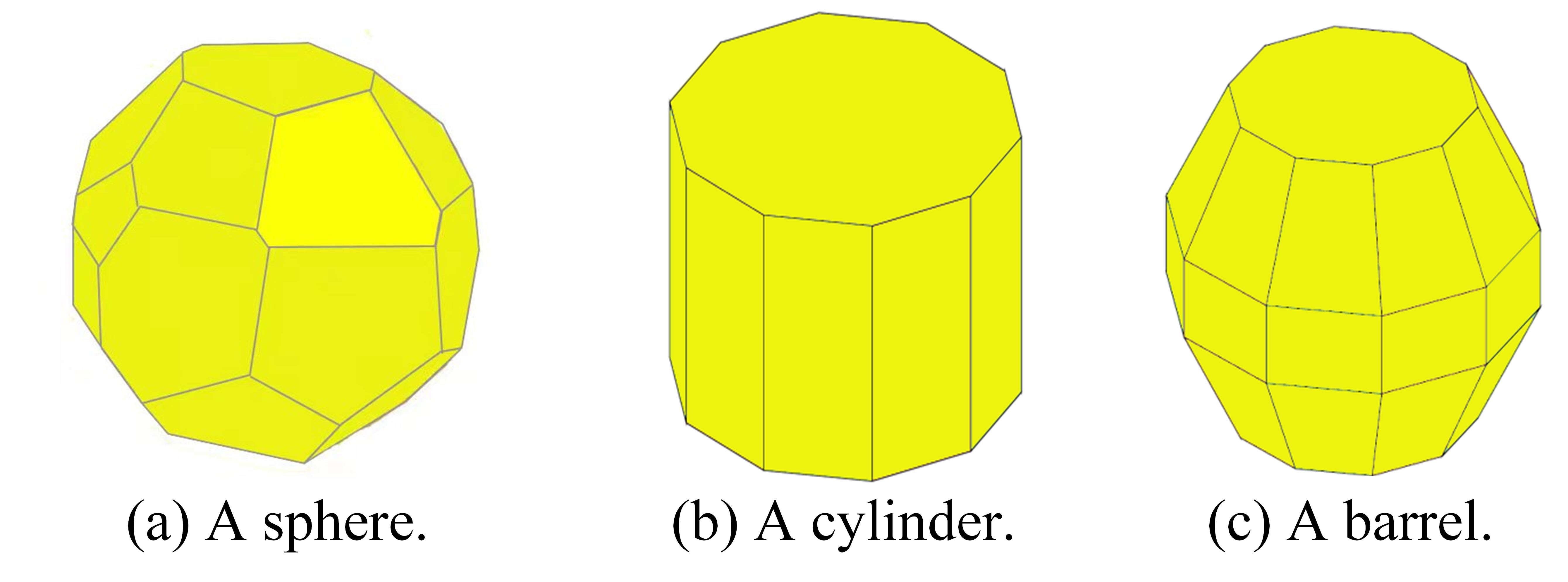}
\caption{
Three examples of convex polyhedron shapes.}
\label{fig:convexpolygons}
\end{figure}

The input to the algorithm is a point cloud obtained from the RGBD data. Artificial scenes contain a great many basic shapes, such as planes, cylinders, and spheres. To make the subsequent algorithms less complicated, the scenes are abstracted to be composed of plane primitives. As shown in Figure~\ref{fig:convexpolygons}, some common objects with curved surfaces can also be abstracted into planes. In the application of planar primitive extraction, the RANSAC algorithm has better comprehensive performance. Based on the algorithm, we adopt an iterative strategy called "fit and remove" to extract each plane primitive in the scene. We compute the convex hull of each primitive regarded as a convex polygon to better understand the topological relationship between pairs of planar primitives. However, the extracted set of planar primitives is not suitable for inferring the scene structure as it only conveys information about local geometric features. Therefore, the planar primitives are converted into a higher-level graph structure that encodes the planar components of the scene and their adjacency relations.

\subsection{Construction of the adjacency graph}
\label{sec:coag}
We convert the extracted set of planar primitives into a semantically richer adjacency graph $G_{adj}=(V,E)$. In this structure, the set of vertices \textit{V} encodes the features of primitives, and \textit{E} represents the adjacency relationships between such primitives. More specifically, \textit{E} is composed of all of the pairs $(v_i,v_j)$ for which the corresponding vertices $v_i,v_j\in V$ are adjacent (with respect to a threshold $\theta_{adj}$). Following the approach described by \cite{mattausch2014object}\cite{mura2016piecewise}\cite{su2021anchor}, we also abstract the scene into a higher-level graph structure composed of planar primitives. However, the neighbor relationship is only the simplest kind of topology information. Therefore, we analyze the intersections of the neighbor plane pairs, especially when the plane pairs include horizontal planes.

\subsection{Detection of pairwise structure}
For the detection of pairwise structure, 3D scenes are abstracted into adjacency graphs composed of plane primitives since they usually contain rich support information. Objects can still be placed on a plane even though there is a small inclination angle between the plane and the ground. We divide the plane primitives in the graph into horizontal and non-horizontal plane primitives in order to infer more topological information in the adjacency graph. It is assumed that when the angle between the primitive and the ground is $<\theta_{angle}$, it is a horizontal primitive. Otherwise it is a non-horizontal primitive. Based on the known ground, \cite{silberman2012indoor}\cite{zhuo2017indoor}\cite{yang2017support} inferred inference support relations. Thus, the ground normal is taken as a priori in our approach.

\begin{figure}[htb]
\centering
\includegraphics[width=.6\linewidth]{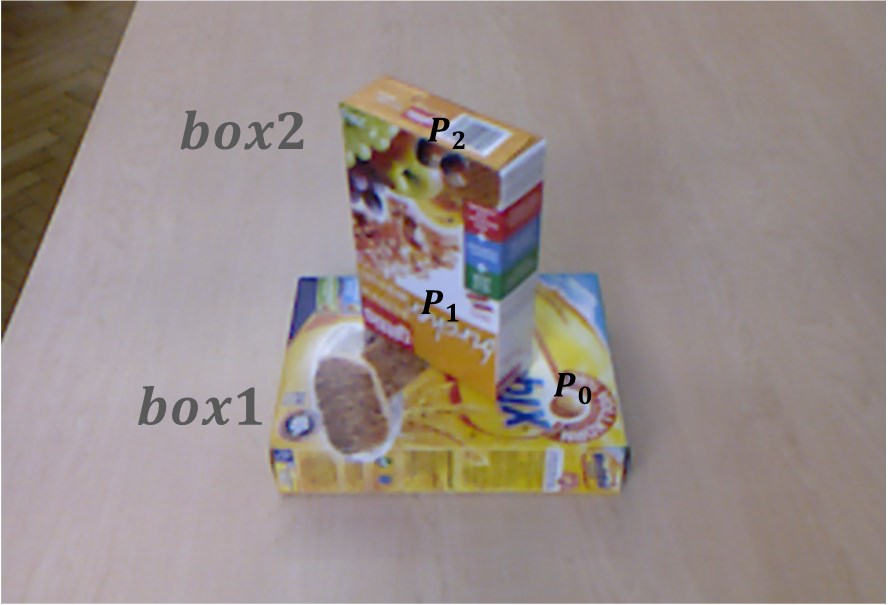}
\caption{
Local support connection and local inner connection. $box_2$ rests on the plane $P_0$ of $box_1$. The spatial configuration of $P_0$ and $P_1$ is considered as a local support relation, while that of $P_1$ and $P_2$ (the planes of $box_2$) is considered as a local inner connection. }
\label{fig:lsclicdef}
\end{figure}

Furthermore, we define the concept of local support connection and local inner connection. As shown in Figure~\ref{fig:lsclicdef}, $P_0$ supports $box_2$ and is a supporting plane of $box_1$. It is considered to support $P_1$ locally where $P_0\in box_1$ and $P_1\in box_2$. This kind of spatial configuration is defined as a local support connection. Meanwhile, the spatial configuration of $P_1$ and $P_2$ (two primitives from $box_2$) is defined as a local inner connection.

\begin{figure}[htb]
\centering
\includegraphics[width=\linewidth]{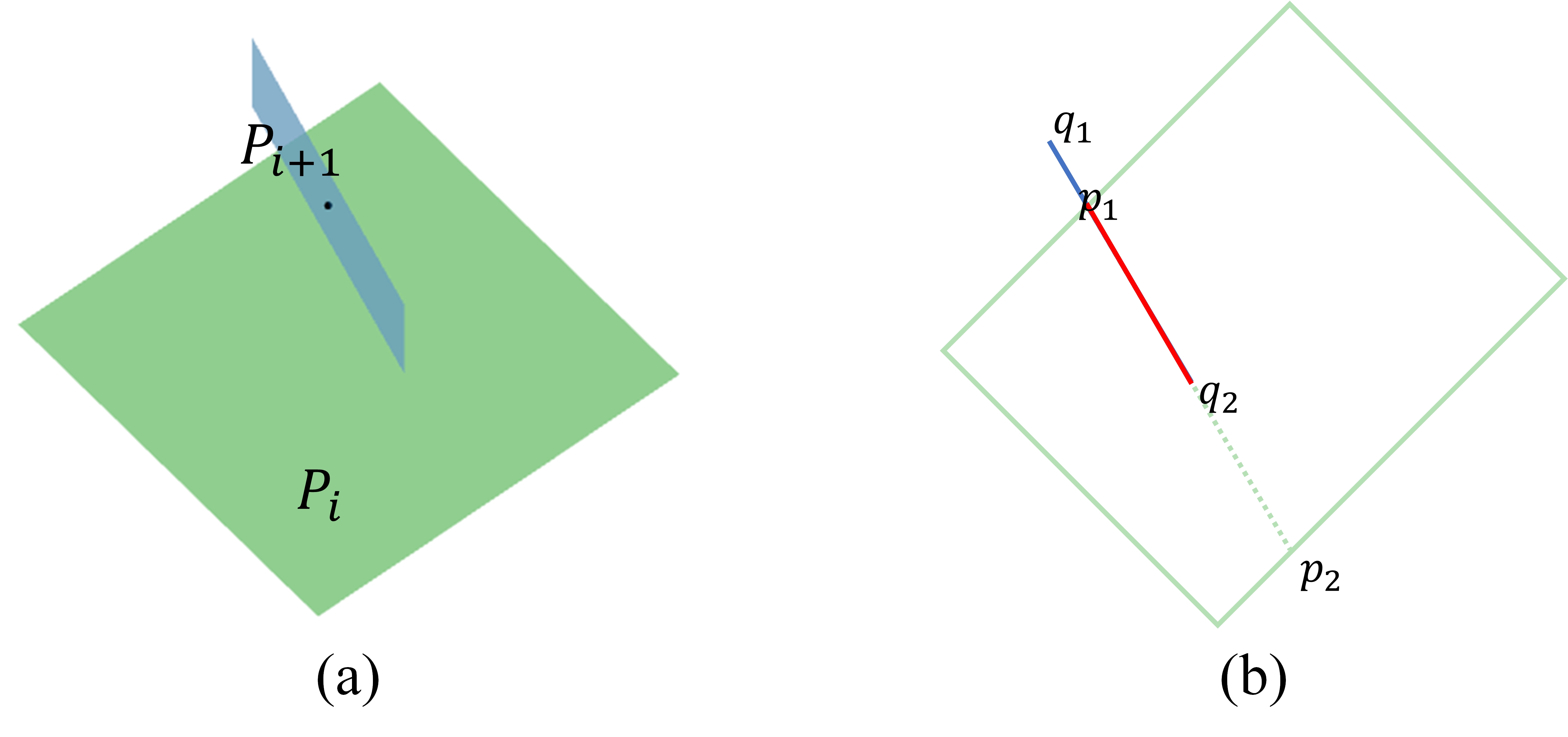}
\caption{
Definition for $\textbf{Ratio}$, which is used to determine whether two neighoring primitives are watertight. (a) An example of a primitive pair configuration. (b) A planar view of (a). In (b), $q_1q_2$ is the bottom edge of plane $P_{i + 1}$ and rests on plane $P_i$. Point $p_1$ and $p_2$ are intersections between line $q_1q_2$ and the edges of plane $P_i$. In this case, $\textbf{Ratio} = min(\frac{p_1q_2}{p_1p_2}, \frac{p_1q_2}{q_1q_2})$. If $\textbf{Ratio} \ge \tau$, two primitives are watertight; otherwise, they are not. Experimentally, $\tau = 0.86$. $\textbf{Ratio}$ for other pairwise spatial configurations in Figure~\ref{fig:spatialpattern} is similar to this case.}
\label{fig:ratiodef}
\end{figure}

\begin{figure*}[tbp]
\centering
\includegraphics[width=\linewidth]{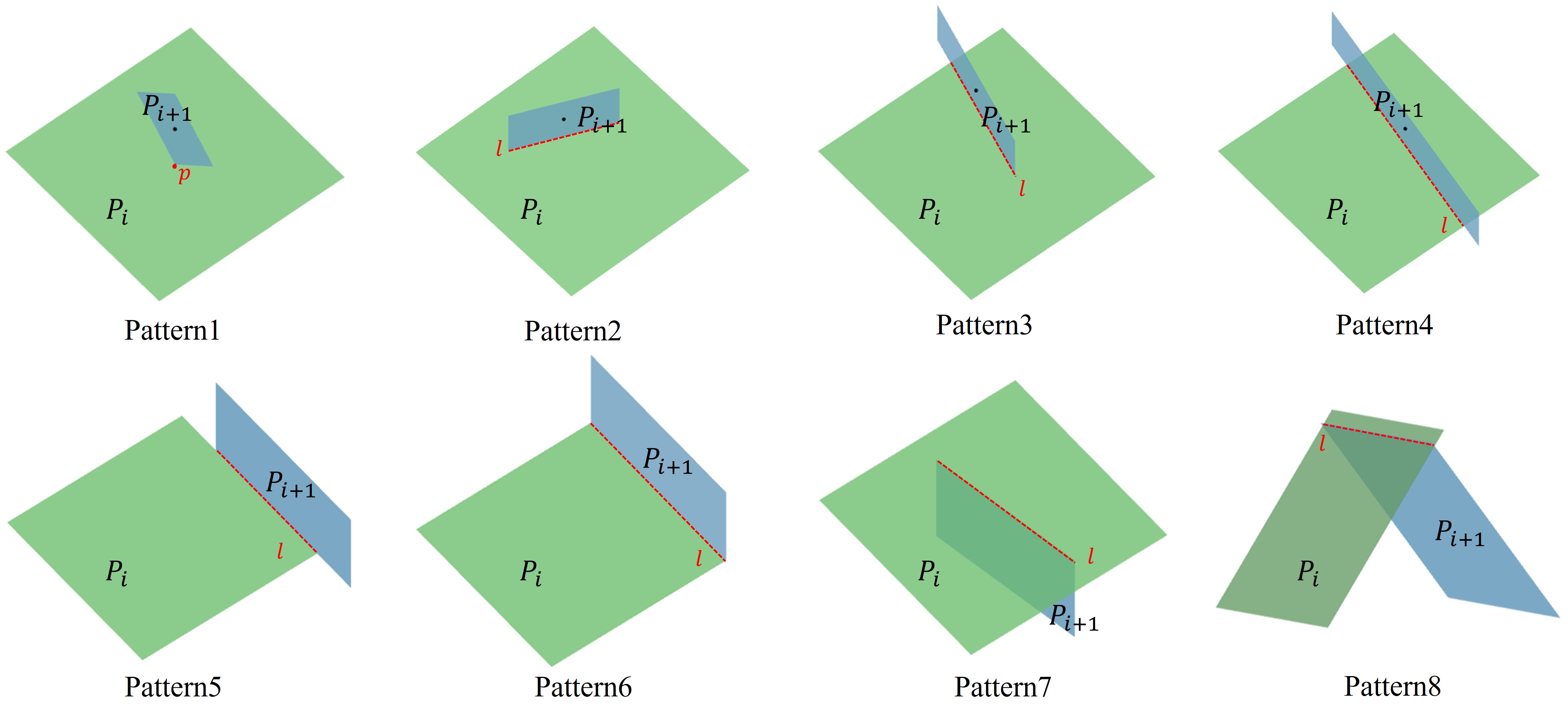}
\caption{%
The eight structural patterns for pairs of neighboring fitting plane primitives are used in our structural detection algorithm.}
\label{fig:spatialpattern}
\end{figure*}

We try to estimate whether there exist local support connections through the spatial configuration of primitive pairs, which is conducive to the subsequent classification of primitives. To label primitives validly, a set of eight spatial configurations are used for encoding. They are denoted as structural patterns and depicted in Figure~\ref{fig:spatialpattern}. \textbf{Pattern1--4} encode four common plane pair configurations that include local support connection between objects. \textbf{Pattern5} and \textbf{Pattern6} have similar spatial configurations, but \textbf{Pattern6} is watertight where the two primitives intersect, while \textbf{Pattern5} is not. Therefore, it is believed that \textbf{Pattern6} belongs to the local inner connection for a single object, and \textbf{Pattern5} belongs to the local support connection between objects. $\textbf{Ratio}$ is defined in Figure~\ref{fig:ratiodef} to distinguish \textbf{Pattern5} from \textbf{Pattern6}. In \textbf{Pattern7}, $P_{i+1}$ is below the horizontal plane $P_i$. $P_{i+1}$ is more inclined to be the inner support of $P_i$ for a single object, so the two planes in \textbf{Pattern7} also belong to the local inner connection. \textbf{Pattern8} captures that the plane pair does not contain a horizontal plane. The plane pair of \textbf{Pattern8} is regarded as a common local inner connection. In particular, the mutual primitive positions for these patterns can be described as follows.
\begin{itemize}
\item[1.] \textbf{Pattern1}: $P_{i+1}$ rests on the horizontal plane $P_i$ and they intersect at point $p$;
\item[2.] \textbf{Pattern2}: $P_{i+1}$ rests on the horizontal plane $P_i$, and the bottom edge $s_{i+1}$ of $P_{i+1}$ intersects $P_i$ and lies on the inside of $P_i$;
\item[3.] \textbf{Pattern3}: $P_{i+1}$ rests on the horizontal plane $P_i$, the bottom edge $s_{i+1}$ of $P_{i+1}$ and $P_i$ are spatially close, and only one intersection exists between the edge $s_{i+1}$ and the edges of $P_i$; 
\item[4.] \textbf{Pattern4}: $P_{i+1}$ rests on the horizontal plane $P_i$, the bottom edge $s_{i+1}$ of $P_{i+1}$ and $P_i$ are spatially close, and two intersections exist between the edge $s_{i+1}$ and the edges of $P_i$; 
\item[5.] \textbf{Pattern5}: an edge of the horizontal plane $P_i$ and $P_{i+1}$ are adjacent but not watertight;
\item[6.] \textbf{Pattern6}: an edge of the horizontal plane $P_i$ and $P_{i+1}$ are adjacent and watertight;
\item[7.] \textbf{Pattern7}: $P_{i+1}$ is under the horizontal plane $P_i$, and they are spatially close;
\item[8.] \textbf{Pattern8}: $P_i$ and $P_{i+1}$ are non-horizontal, and they are spatially close.     
\end{itemize}

\begin{figure*}[tbp]
\centering
\includegraphics[width=0.75\linewidth]{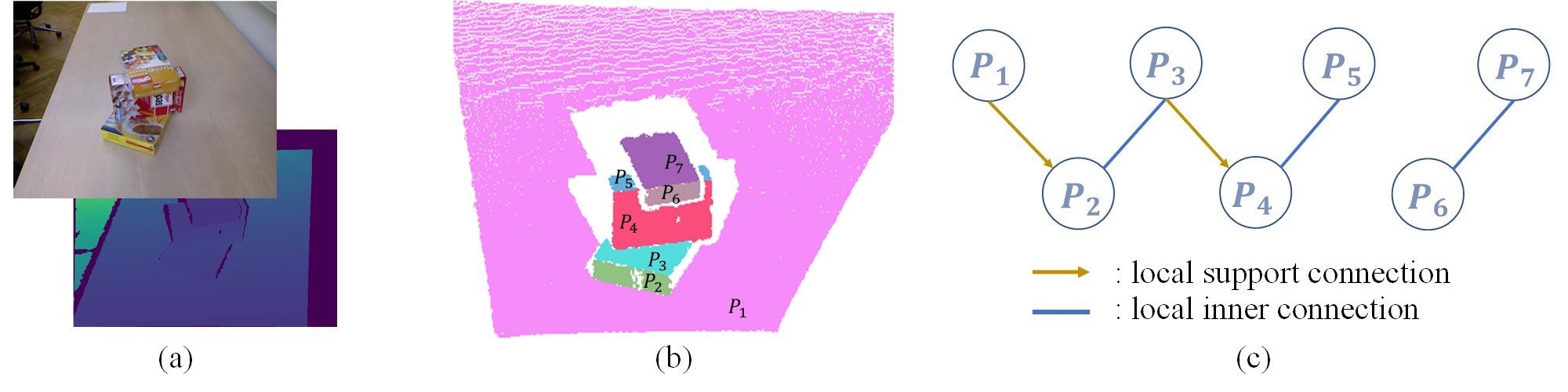}
\caption{
An example of the detection of neighboring primitives spatial configuration. (a) RGB and depth images from RGBD data. (b) Abstraction of the scene to planar primitives. (c) Result of detection of pairwise spatial configuration. All of the spatial configurations of neighboring primitives are divided into two categories: local support connection and local inner connection.} 
\label{fig:empofconfiguration}
\end{figure*}

As shown in Figure~\ref{fig:spatialpattern}, \textbf{Pattern1--5} are considered as local support connections between objects, and \textbf{Pattern6--8} are considered as local inner connections for a single object. Figure~\ref{fig:empofconfiguration} shows an example of the spatial configuration detection results of primitives in the scene graph.

\section{Primitive Classification}
\label{sec:primitive_classification}
In this section, we label the primitives in the scene based on the two kinds of pairwise structures in the adjacency graph model: local support connection between objects and local inner connection for a single object. Further, we transform primitive classification into the combinatorial optimization problem. 

We assume that there are $N$ nodes in graph $G=(V, E)$. To simplify the problem of initial label assignment, we establish a primary hypothesis that there exist $N$ labels $L=\{l_1,l_2,\ldots,l_N\}$ . Our goal is to determine the optimal assigned label $L^*=\{L^*_v|L^*_v \in L, v \in V\}$ to maximize the energy function.

\begin{equation}
\mathop{argmax}_{L^*} E(L)  = E_{data}(L) + E_{smooth}(L)
\label{eq1}
\end{equation}

The objective function Eq.~\ref{eq1} consists of a data term representing the weight of the coarse label assignments of primitives and a smooth term indicating the weight of the coherent label assignments of adjacent primitives. We introduce a binary indicator matrix $X_{N*N}$, in which each element is denoted as $x_{v_i \leftarrow l_j}\in\{0,1\}$. And $x_{v_i \leftarrow l_j} =1$ indicates that label $l_j$ is assigned to primitive $v_i$; otherwise, label $l_j$ is not assigned. Therefore, the objective function is rewritten as follows:

\begin{equation}
\begin{split}
&\mathop{argmax}\limits_{X^*}E(X)  =  \sum\limits_{x_{v_i\leftarrow l_j} \in X, v_i \in V, l_j \in L}x_{v_i\leftarrow l_j} \cdot D(v_i,l_j) \\
& \quad \quad  +  \frac{1}{2} \sum\limits_{j = 1}^{N_{labels}} \sum\limits_{m=1}^{N_{nodes}} \sum\limits_{n=1}^{N_{nodes}} x_{v_m \leftarrow l_j} \cdot x_{v_n\leftarrow l_j} \cdot F(v_m,v_n)  \\
&\quad subject\ to \quad \sum\limits_{j=1}^{N_{labels}} x_{v_i\leftarrow l_j} = 1, \forall v \in V,\forall l \in L 
\end{split}
\label{eq2}
\end{equation}
where $N_{labels}=N_{nodes}=N$, and $\sum\limits_{j=1}^{N_{labels}} x_{v_i\leftarrow l_j} = 1, \forall v \in V,\forall l \in L$  establishes the constraint that each primitive can only be assigned one label.

\textbf{Data term.} $E_{data}$ is the sum of the label weights of all the nodes, where each label weight represents encouragement for assigning label $l_j$ to primitive $v_i$:

\begin{equation}
E_{data}(X)=  \sum\limits_{x_{v_i\leftarrow l_j} \in X, v_i \in V, l_j \in L}x_{v_i\leftarrow l_j} \cdot D(v_i,l_j)
\label{dataterm}
\end{equation}

To derive the data term, we set the coarse weights of labels assigned to primitives in terms of the detection of pairwise spatial structure. Neighboring primitives belonging to local support connection are assigned different labels. Primitives belonging to local inner connection are assigned the same label. The data term $D(v_i, l_j)$ is expressed as follows:

\begin{footnotesize}
\begin{equation}
D = 
\begin{cases}
D_{v_i \leftarrow l_i}=d_0, &\\
D_{v_i\leftarrow l_j}=D_{v_j\leftarrow l_i}=-\infty, &if \ PT(v_i, v_j) \in\{1,2,3, 4\}\\ &\\

D_{v_i\leftarrow l_i} = d_0 \cdot RTO, &\\ 
D_{p_i\leftarrow l_j}=D_{p_j\leftarrow l_i} = - \infty, &if \ PT(v_i, v_j) \in \{5\} \\ &\\
D_{v_i\leftarrow l_j} = D_{v_j \leftarrow l_i} = d_0 \cdot RTO, &if \ PT(v_i, v_j) \in \{6, 7\} \\ & \\ 
D_{v_i\leftarrow l_j} = D_{v_j\leftarrow l_i}=d_0 \cdot \mathcal{F}[\cdot], &if \ PT(v_i, v_j)\in \{8\} \\ & \\
D_{v_i\leftarrow l_j} = D_{v_j\leftarrow l_i}=0, &otherwise
\end{cases}
\label{dfunc}
\end{equation}
\end{footnotesize}
where $PT(.,.)$ represents the spatial configuration of two neighboring primitives as defined in Figure~\ref{fig:spatialpattern}, $RTO(.,.)$ computes the $\textbf{Ratio}$ between two neighoring primitives, which is described in details in Figure~\ref{fig:ratiodef}. The $\textbf{Ratio}$ is used to measure whether the two primitives are watertight. $d_0$ is a positive constant and $\mathcal{F}[\cdot]$ is denoted as follows:
\begin{equation}
    \mathcal{F}[\cdot] = 
    \begin{cases}
    RTO, &if \ \delta \le RTO \le 1 \\
    - RTO, & if \ 0 \le RTO < \delta
    \end{cases}
    \label{eq:f-rto}
\end{equation}
where $\delta$ is experimentally set to 0.2.

\textbf{Smooth term.} The effect of $E_{data}$ is balanced by the smoothness energy $E_{smooth}$, which aims at encouraging edges involved in local inner connection and penalizing edges involved in local support connection. $E_{smooth}$ computes the sum of weights of edges between neighboring primitives assigned the same labels.

\begin{equation}
\begin{split}
&E_{smooth}(X) = \\
&\quad \quad \quad \quad \frac{1}{2} \sum\limits_{j = 1}^{N_{labels}} \sum\limits_{m=1}^{N_{nodes}} \sum\limits_{n=1}^{N_{nodes}} x_{v_m \leftarrow l_j} \cdot x_{v_n\leftarrow l_j} \cdot F(v_m,v_n)
\label{smoothterm}
\end{split}
\end{equation}
where $F(.,.)$ represents the connection weight function of the primitive pair, and $v_m,v_n\in V$ . If the primitive pair belongs to the support connection, to penalize the assignment of the same label, the connection weight is set to a negative number; if it belongs to the inner connection, the weight is set to a positive number. The connection weight function $F(.,.)$ is expressed as follows:

\begin{footnotesize}
\begin{equation}
F =
\begin{cases}
(w_d \cdot DIST + w_r \cdot RTO) \cdot f_0, & if\  PT(v_i, v_j)\in \{6, 7\} \\
(w_d \cdot DIST + w_r \cdot \mathcal{F}[\cdot]) \cdot f_0, & if\ PT(v_i, v_j) \in \{8\} \\
-f_1, & if\  PT(v_i, v_j) \in \{1, 2, 3, 4,5\} \\
0, & otherwise
\end{cases}
\label{ffunc}
\end{equation}
\end{footnotesize}
where $PT(.,.)$ represents the spatial configuration of two neighboring primitives as described in Figure~\ref{fig:spatialpattern}, $DIST(.,.)$ indicates the distance of the primitive pair normalized by $\theta_{adj}$(denoted in~\ref{sec:coag} ). $RTO(.,.)$ measures the watertight between two primitives, and $\mathcal{F}[\cdot]$ is defined in Eq.~\ref{eq:f-rto}. Additionally, $w_d = 0.3$, $w_r=0.7$ and $d_0 = f_0 \ll f_1$.

The objective equation Eq.~\ref{eq2} is a linearly constrained unary quadratic integer programming problem, which is an NP complete problem. Gurobi Optimizer is used to solve this problem.

\section{Support Relations Inference and Hierarchy Graph Construction}
\label{sec:support_relations_inference}
Combinatorial optimization is used to assign labels to primitives in the adjacency graph, and each label is considered as an object. In this situation, the scene graph contains both label information of objects and local support information between objects. Further, the support relation between objects is inferred according to local support information and label information. The following assumptions are applied in our model: (a) the other primitives are above the ground, which means the ground is the lowest planar primitive in the scene; (b) the object containing the ground requires no support; and (c) each object, with the exception of the object containing ground, is supported by other detected objects or invisible objects.

In the scene, $K$ objects are detected and denoted as $C_K=\{C_1,\ldots,C_K\}$. Also, we introduce an invisible object $C_{K+1}$. $C_j$, with the exception of the object containing ground, is supported by $C_i$ and denoted as $SUPP(C_i,C_j)=1, 1\le i,j\ \le K$; $C_i$ supported by invisible objects is denoted as $SUPP(C_{K+1},C_i)=1, 1\ \le i\ \le K$; and the object containing ground requires no support and is denoted as $SUPP(C_{K + 1}, C_{ground})=1$. 

\begin{figure*}[tbp]
\centering
\includegraphics[width=0.75\linewidth]{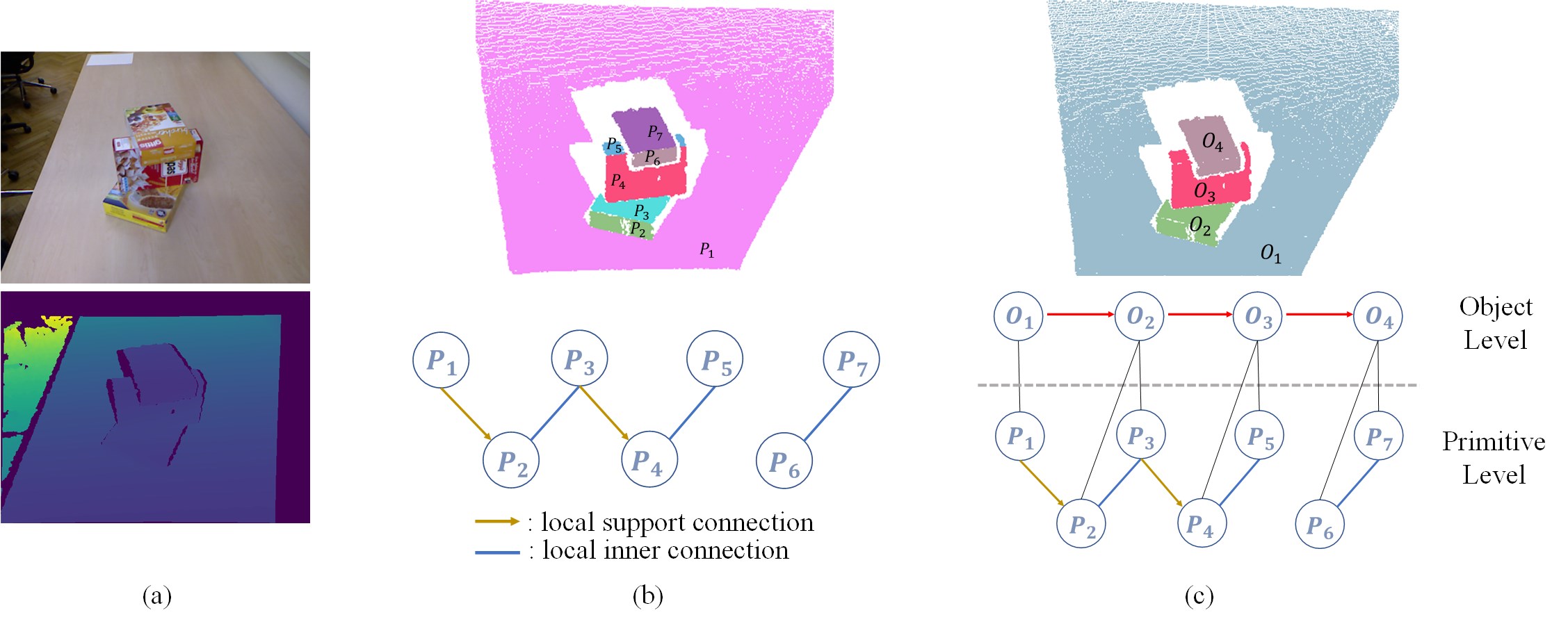}
\caption{
An example of support relations inference and scene hierarchy graph construction. (a) The RGB and depth images from RGBD data. (b) The scene is abstracted to an adjacency graph with planar primitives, and the edges of it are divided into two categories: local support connection and local inner connection. (c) Scene hierarchy graph. A root node is introduced and indicates an invisible object supporting the object containing ground plane and objects without detected supporters in the scene. However, in this scene, the root is omitted, because only the object $O_1$ has no supporter.}
\label{fig:empofsupportinference}
\end{figure*}

\begin{algorithm}
\caption{Support Relations Inference and Scene Hierarchy Graph Construction.}
\label{alg:A}
\begin{algorithmic}
\STATE \textbf{for} i = 1 : $K$ \textbf{do}
\STATE \quad \textbf{if} $C_i$ contains ground:
\STATE \quad \quad $SUPP(C_{K+1},C_i) = 1$
\STATE \quad \quad \textbf{continue}
\STATE \quad \textbf{for} j = i + 1 : $K$ \textbf{do}
\STATE \quad \quad \textbf{if} $LocalSupport(C_i, C_j) = 1$ :
\STATE \quad \quad \quad $SUPP(C_i, C_j) = 1$
\STATE \quad \quad \quad \textbf{continue}
\STATE \quad \quad \textbf{if} $LocalSupport(C_j, C_i) = 1$:
\STATE \quad \quad \quad $SUPP(C_i, C_j) = 1$
\STATE \quad \textbf{end for}
\STATE \textbf{end for}
\STATE
\STATE \textbf{for} i = 1 : $K$ \textbf{do}
\STATE \quad \textbf{if} $C_i$ contains ground:
\STATE \quad \quad \textbf{continue}
\STATE \quad \textbf{for} j = 1: $K$ \textbf{do}
\STATE \quad \quad \textbf{if} $SUPP(C_j, C_i) \neq 1$:
\STATE \quad \quad \quad \textbf{if} $GlobalSupport(C_j, C_i)$ = 1:
\STATE \quad \quad \quad \quad $SUPP(C_j, C_i) = 1$
\STATE \quad \textbf{end for}
\STATE \textbf{end for}
\STATE
\STATE \textbf{for} i = 1 : $K$ \textbf{do}
\STATE \quad \textbf{if} no object supports $C_i$:
\STATE \quad \quad $SUPP(C_{K+1}, C_i) = 1$
\STATE \textbf{end for}
\end{algorithmic}
\end{algorithm}

The scene hierarchy graph is constructed by using support relations as described in Algorithm~\ref{alg:A}, and an example is shown in  Figure~\ref{fig:empofsupportinference}. Support relations inference consists of two parts: local support relations inference and global support relations inference. 

First, the support relations are inferred from primitive level to object level with the result of neighboring pairwise spatial configuration. The function $LocalSupport(C_i, C_j)$ is defined, where $1\le i,j \le K$. If $LocalSupport(C_i, C_j) = 1$, it means that one planar primitive of object $C_i$ supports one or more primtives of $C_j$ and thus $SUPP(C_i, C_j) = 1$. Note that $LocalSupport(C_i, C_j) \neq LocalSupport(C_j, C_i)$. In Figure~\ref{fig:empofsupportinference}, it is inferred that object $O_1$ supports $O_2$ because primitive $P_1$ from $O_1$ supports $P_2$ of $O_2$ locally, which is denoted as $SUPP(O_1, O_2) = 1$. In the same way, $SUPP(O_2, O_3) = 1$.

Although some local support relations in the scene are mined by the pairwise configuration detection module, the occlusion in the single-view RGBD data will still cause the local support relations between objects to be undetectable. Global support relations inference is proposed to infer the left support relations in the scene. The function $GlobalSupport(C_i, C_j)$ is defined, where $1 \le i, j \le K$ and $LocalSupport(C_i, C_j) \ne 1$. If $GlobalSupport(C_i, C_j) = 1$, it means that there exists a planar primitive $P^*$ from $C_i$ supporting the $bbox$ \cite{jia20133d} of the object $O_j$ optimally, and $SUPP(C_i, C_j) = 1$. For example, in Figure~\ref{fig:empofsupportinference}, it is inferred that $O_3$ supports $O_4$ with the global support relations inference because the planar primitive $P_5$ from Object $O_3$ supports the $bbox$ of $O_4$.

Finally, an invisible object is introduced to support the object containing ground and the objects without detected supporters in the scene. In the hierarchy graph, the invisible object is denoted as the root, which ensures that other nodes can be traversed. Only if  the object containing ground has no supporter in the scene, the root is omitted.

Note that the algorithm introduces a root as a hidden node, which connects the object containing the ground and objects without detected supporters in the scene. Unlike \cite{silberman2012indoor}\cite{yang2017support}\cite{zhuo2017indoor}, we consider the case where an object is supported by one or more objects.

\begin{figure*}[tbp]
\centering
\includegraphics[width=\linewidth]{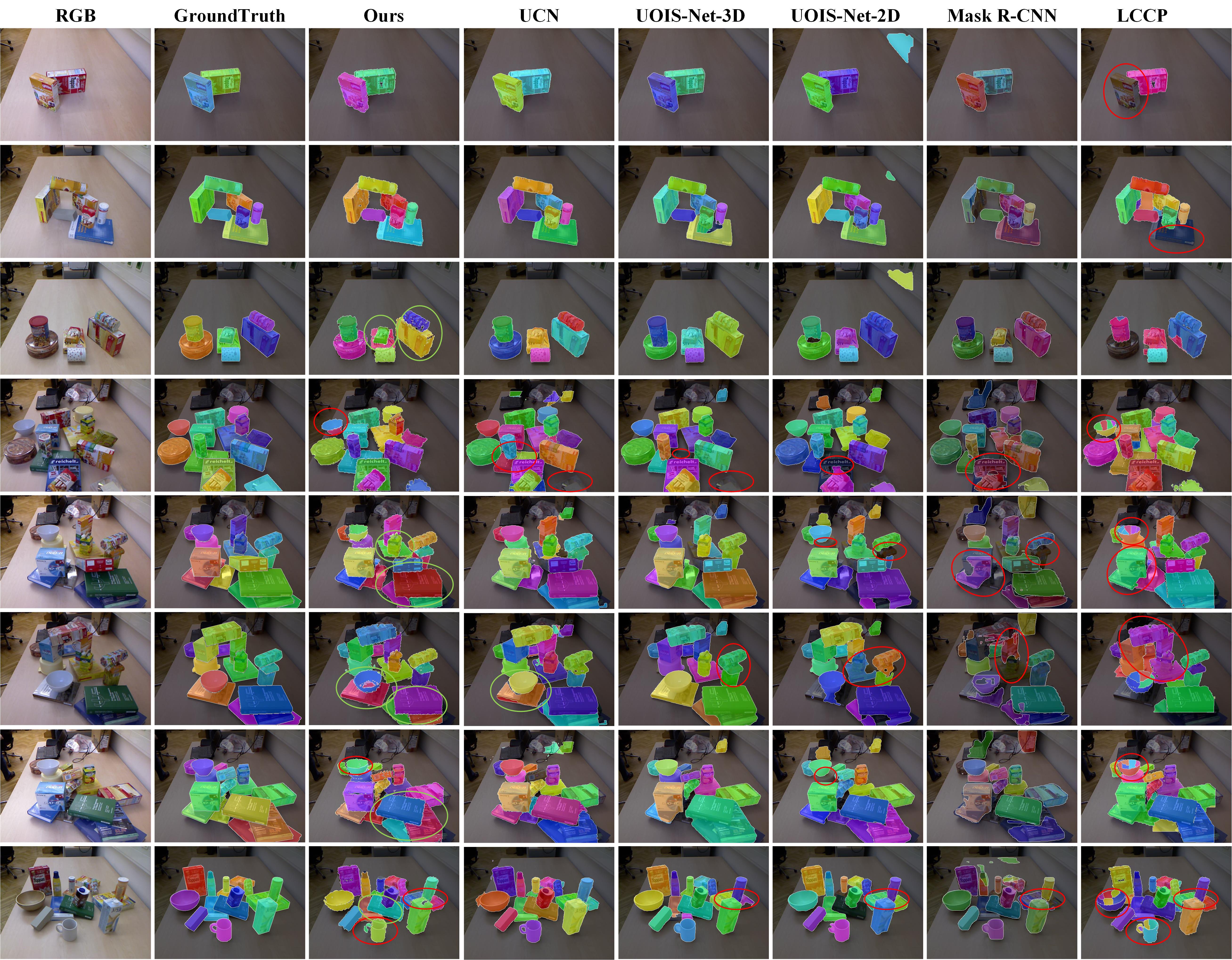}
\caption{
Comparison of our method with baselines, UCN \cite{xiang2021learning}, UOIS-Net-3D\cite{xie2021unseen}, UOIS-Net-2D\cite{xie2020best}, Mask R-CNN\cite{he2017mask} and LCCP\cite{christoph2014object}. The first column is RGB images and the second column is the ground truth of segmentation. The third and the seventh columns are the experimental results of our algorithm and LCCP, both operating on depth images. The fourth and the fifth columns are UOIS-Net-3D and UOIS-Net-2D, both of their input are RGBD images. The input to the sixth column Mask R-CNN is RGB images. Green circles indicate the better segmentation results compared with the other algorithms, while red indicate errors. The more green circles are, the better segmentation performance is. Our propsed method provides accurate masks in comparison to all of the baselines.}
\label{fig:segmentationsres}
\end{figure*}

\begin{figure*}[tbp]
\centering
\includegraphics[width=\linewidth]{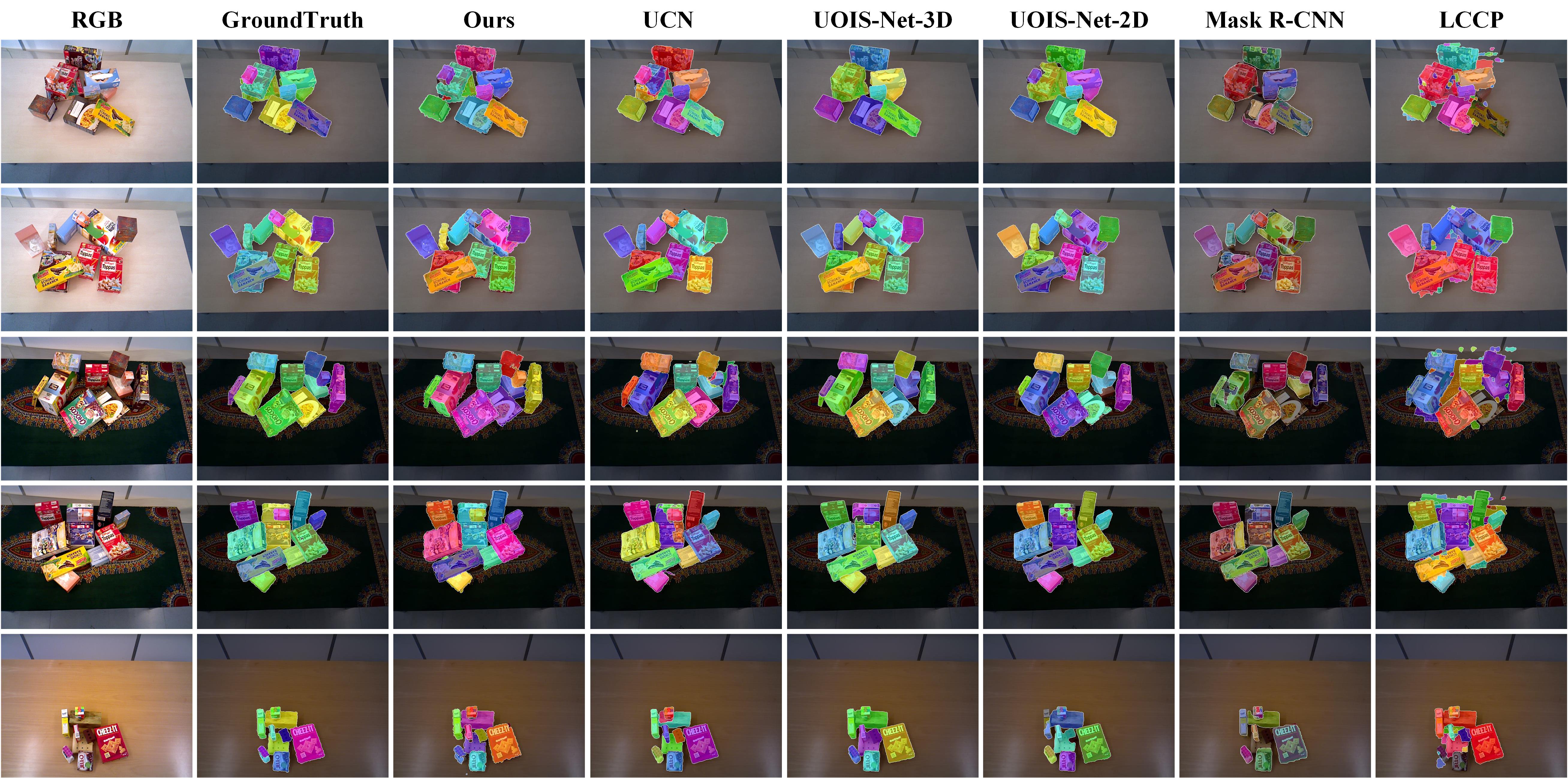}
\caption{
Comparison of our method with baselines, UCN \cite{xiang2021learning}, UOIS-Net-3D\cite{xie2021unseen}, UOIS-Net-2D\cite{xie2020best}, Mask R-CNN\cite{he2017mask} and LCCP\cite{christoph2014object}. The first column is RGB images and the second column is the ground truth of segmentation. The third and the seventh columns are the experimental results of our algorithm and LCCP, both operating on depth images. The fourth and the fifth columns are UOIS-Net-3D and UOIS-Net-2D, both of their input are RGBD images. The input to the sixth column Mask R-CNN is RGB images. Green circles indicate the better segmentation results compared with the other algorithms, while red indicate errors. The more green circles are, the better segmentation performance is. Our propsed method provides accurate masks in comparison to all of the baselines.}
\label{fig:ocid-segmentationsres}
\end{figure*}

\section{Experiments and Results}
\label{sec:experiments}
In this section, we quantitatively evaluate our algorithm for its performance in pairwise spatial configuration detection, primitive classification, and scene graph support inference based on the OSD dataset \cite{richtsfeld2012segmentation} and the OCID dataset \cite{suchi2019easylabel}. OSD, a small dataset with rich support relations and high-quality human annotation, contains a total of 111 RGBD images of stacked and occluding objects on tables. In a scene, the more complex the support relations are, the more difficult the segmentation is. We mainly examine the performance of our method on scenes with complex support relations in the OSD dataset. Our algorithm is implemented in C++ on a 
laptop equipped with a 2.3 GHz processor and 16GB of memory.

\subsection{Evaluating Primitives Classification and Instance Segmentation}

Calculating the accuracy of local support connections and local inner connections in all test scenes, we evaluate the neighboring primitive configuration detection model because it affects the subsequent primitive classification and support relations inference. As shown in Table~\ref{tb1}, local support connections and local inner connections both have high accuracy which ensures the accuracy of the subsequent correct primitive classification. 

\begin{table}[htb]
\caption{Accuracy of neighboring primitives structure detection and primitive classification.}
\label{tb1}
\centering
\begin{tabular}{c c c c}
\toprule
&local support&local inner &primitive\\
&connection &connection &classification \\
\hline
Accuracy & 0.83 & 0.88 & 0.85 \\
\bottomrule
\end{tabular}
\end{table}

We also evaluate the performance of primitive classification by comparing our algorithm with UOIS-Net-3D\cite{xie2021unseen}, UOIS-Net-2D\cite{xie2020best}, Mask R-CNN\cite{he2017mask} and LCCP\cite{christoph2014object}. Figure~\ref{fig:segmentationsres} shows some qualitative results of our method and baselines. We can see that our proposed method performs best compared with baselines, especially when objects are stacked together with complex support relations. Our algorithm is also robust to color similarity while UOIS-Net-3D, UOIS-Net-2D and Mask R-CNN do not perform well. However, as bowls and cups are spatially divided into two or more parts due to occlusion in point cloud, our algorithm does not have a great performance, which is the same as LCCP. UOIS-Net-3D and UOIS-Net-2D perform second only to our algorithm, especially in the scenes containing complex support relations and similar colors between objects. And when objects with similar colors are stacked together, Mask R-CNN trained on RGB images often undersegments them. LCCP, based on concavity and convexity, has the worst performance among all baselines.

Follow \cite{xie2021unseen}, the precision/recall/F-measure (P/R/F) metrics are used to further evaluate the object segmentation performance of our algorithm and baselines. They are computed by
$$
P = \frac{\sum_i|c_i \cap g(c_i)|}{\sum_i|c_i|}, R=\frac{\sum_i|c_i \cap g(c_i)|}{\sum_j|g_j|}, F = \frac{2PR}{P+R},
$$
where $c_i$ denotes the set of pixels belonging to predicted object $i$, $g(c_i)$ indicates the set of pixels of the matched ground truth object $c_i$, and $g_j$ is the set of pixels for ground truth object $j$. Overlap P/R/F metrics are denoted above and the true positives are counted by pixel overlap of the whole object. To evaluate how sharp the predicted boundary matches against the ground truth  boundary\cite{xie2021unseen}, Boundary introduces P/R/F metrics, where the true positives are counted by pixel overlap of two boundaries from the predicted result and the ground truth. See \cite{xie2021unseen} for more details.

Table~\ref{seg-PRF} shows the results of our method and baselines. As we can see from the table, our method achieves the best Overlap P/R/F metrics. Although we obtain the best Boundary Recall metric, UOIS-Net-3D performs the best on Boundary P/R/F metrics. The main factor of this is that cropping occurs when the RGB and depth are aligned to generate the point cloud, as shown from the fifth row to the seventh row in Figure 8. Another one may be the noise in depth images. Overall, our method performs the best compared with baselines. 

\begin{figure*}[tbp]
\centering
\includegraphics[width=\linewidth]{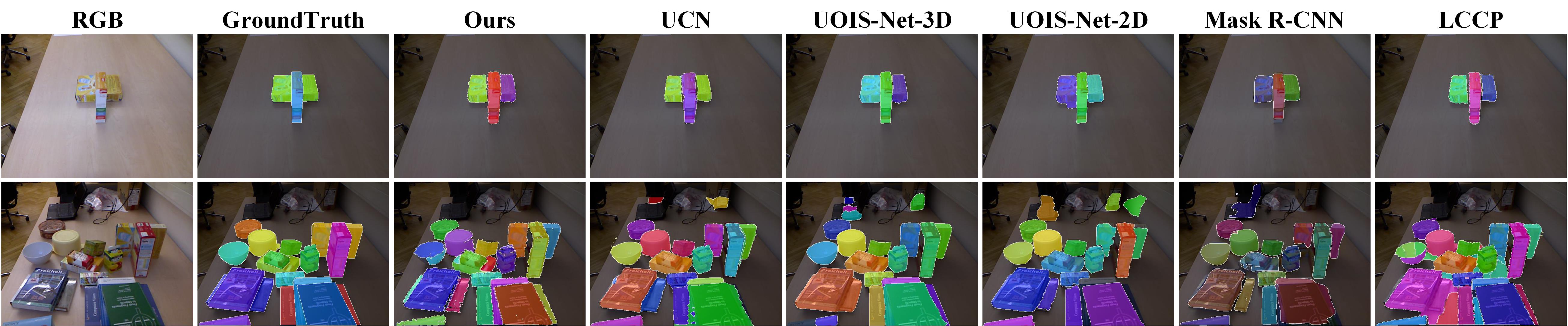}
\caption{Failure cases of segmentation from our method and baselines.  The first column is RGB images  and the second is the ground truth of segmentation. The third to the seventh columns are our method, UCN \cite{xiang2021learning}, UOIS-Net-3D\cite{xie2021unseen}, UOIS-Net-2D\cite{xie2020best}, Mask R-CNN\cite{he2017mask}, and LCCP\cite{christoph2014object}, respectively.
}
\label{fig:failurecases}
\end{figure*}

Figure~\ref{fig:failurecases} shows two failure cases of our method and baselines because of the over-segmentation due to occlusion. As we can see, the surfaces of several boxes and books are divided into two or more parts, especially when objects are stacked together with complex support relations. Future works could investigate how to fix these failures from multi-view images.

\begin{table}
    \caption{Comparison with baselines on OSD dataset. \textcolor{red}{Red} indicates the best performance. All P/R/F measures are shown in the range[0, 100]($P/R/F \times 100$).}
    \label{seg-PRF}
	\centering
	\begin{tabular}{cccccccc}
		\toprule
		\multirow{2}{*}{\textbf{Method}} &
		\multirow{2}{*}{\textbf{Input}} &
		\multicolumn{3}{c}{\textbf{Overlap} }& \multicolumn{3}{c}{\textbf{Boundary} }\\
		 & & P & R & F & P & R & F\\
		\hline
		LCCP\cite{christoph2014object}       & Depth & 60.5 & 62.0 &  61.1  & 45.5 & 51.2 & 48.0  \\
		Mask R-CNN\cite{he2017mask} & RGB   & 74.2 & 77.4 &  74.8  & 53.8 & 57.9 & 54.8 \\
		UOIS-Net-2D\cite{xie2020best}    & RGBD &74.8 & 82.1 &  77.3  & 64.8 & 67.3 & 64.8 \\
	    UOIS-Net-3D\cite{xie2021unseen}    & RGBD & 72.3 & 84.0 &  77.1  & \textcolor{red}{66.5} & 71.3 & \textcolor{red}{67.7} \\
            UCN \cite{xiang2021learning} & RGBD & 72.9 & 89.2 & 78.3 & 56.7 & 71.6 & 62.3 \\
		Ours       & Depth  & \textcolor{red}{78.3} & \textcolor{red}{91.5} &  \textcolor{red}{83.6}  & 61.1 & \textcolor{red}{76.7} & 67.4 \\
		\bottomrule
	\end{tabular}
\end{table}

\begin{table}
    \caption{Comparison with baselines on OCID dataset. \textcolor{red}{Red} indicates the best performance. All P/R/F measures are shown in the range[0, 100]($P/R/F \times 100$).}
    \label{ocid-seg-PRF}
	\centering
	\begin{tabular}{cccccccc}
		\toprule
		\multirow{2}{*}{\textbf{Method}} &
		\multirow{2}{*}{\textbf{Input}} &
		\multicolumn{3}{c}{\textbf{Overlap} }& \multicolumn{3}{c}{\textbf{Boundary} }\\
		 & & P & R & F & P & R & F\\
		\hline
		LCCP\cite{christoph2014object}    & Depth & 60.5 & 64.9 &  62.5  & 37.2 & 50.6 & 42.7  \\
		Mask R-CNN\cite{he2017mask}       & RGB   & 88.9 & 74.6 &  81.0  & 58.5 & 51.6 & 54.7 \\
		UOIS-Net-2D\cite{xie2020best}     & RGBD  & 93.9 & 89.4 &  91.1  & 81.6 & 74.0 & 77.6 \\
	    UOIS-Net-3D\cite{xie2021unseen}   & RGBD  & 91.0 & 90.9 &  91.2  & \textcolor{red}{83.5} & 77.4 & \textcolor{red}{80.1} \\
            UCN \cite{xiang2021learning}     & RGBD & \textcolor{red}{91.7} & 91.3 & \textcolor{red}{91.5} & 74.7 & 81.9 & 77.9 \\
		Ours                              & Depth & 90.3 & \textcolor{red}{92.7} & \textcolor{red}{91.5}  & 82.2 & \textcolor{red}{84.5} & 79.5 \\
		\bottomrule
	\end{tabular}
\end{table}

\subsection{Evaluating Scene Hierarchy Graph}

\begin{figure}[tbp]
\centering
\includegraphics[width=\linewidth]{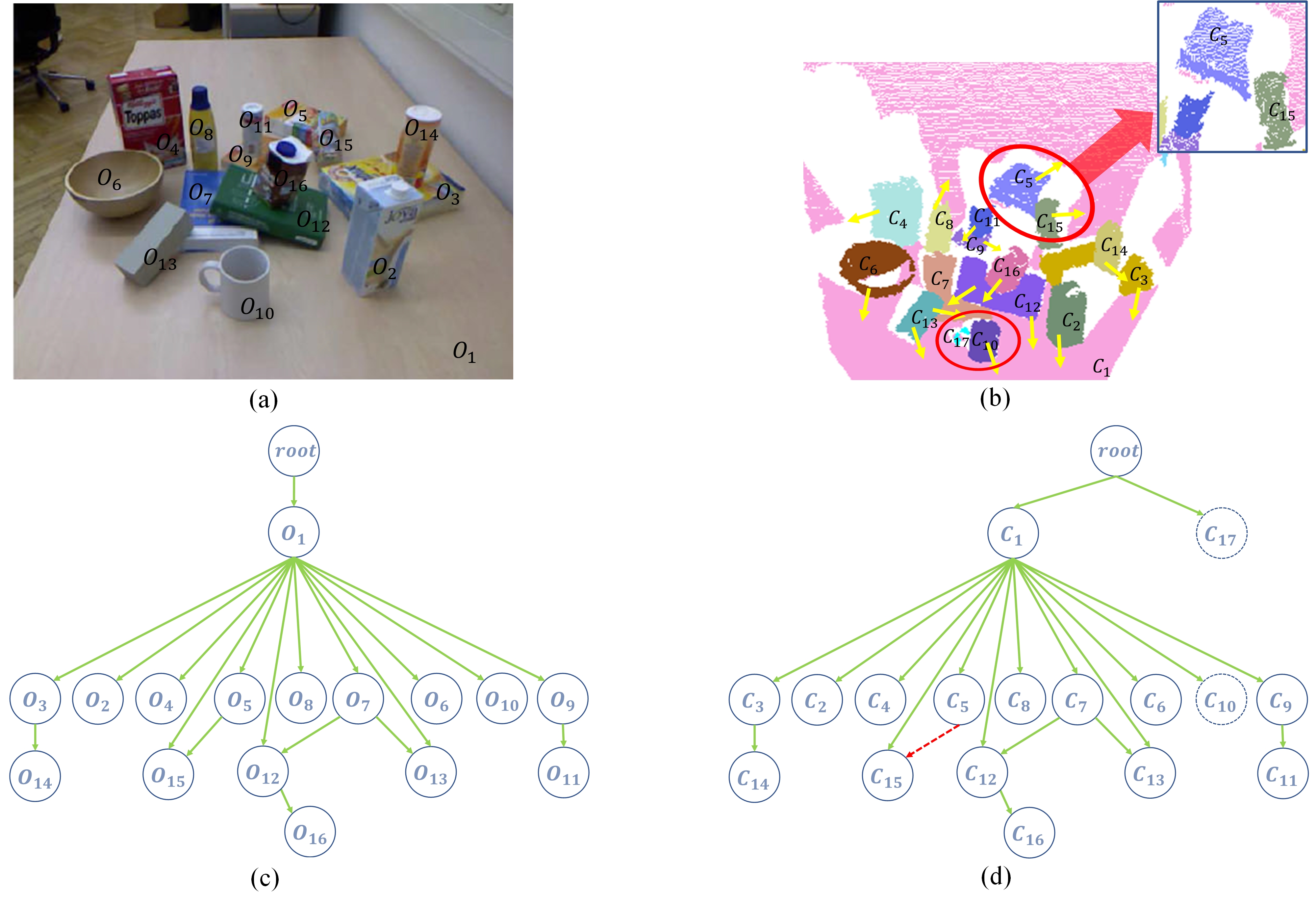}
\caption{
An example of a constructed scene graph. (a) RGB images of the scene in which 16 objects are labeled. (b) The result of segmentation and support relations inference with our method. Errors are circled in red. (c) The ground truth of the scene graph for object level. (d) The result of our scene graph construction. Red dashed arrows in the graph indicate support relations not detected in the scene. In (b), the cup in the scene is incorrectly divided into objects $C_{10}$ and $C_{17}$. Additionally, the support relation between $C_5$ and $C_{15}$ is not detected due to the occlusion. Correspondingly, in the scene graph, $C_{17}$ is supported by an invisible object, which is connected by a root. It lacks a directed edge from $C_5$ to $C_{15}$, too.}
\label{fig:hgc}
\end{figure}

\begin{figure}[tbp]
\centering
\includegraphics[width=\linewidth]{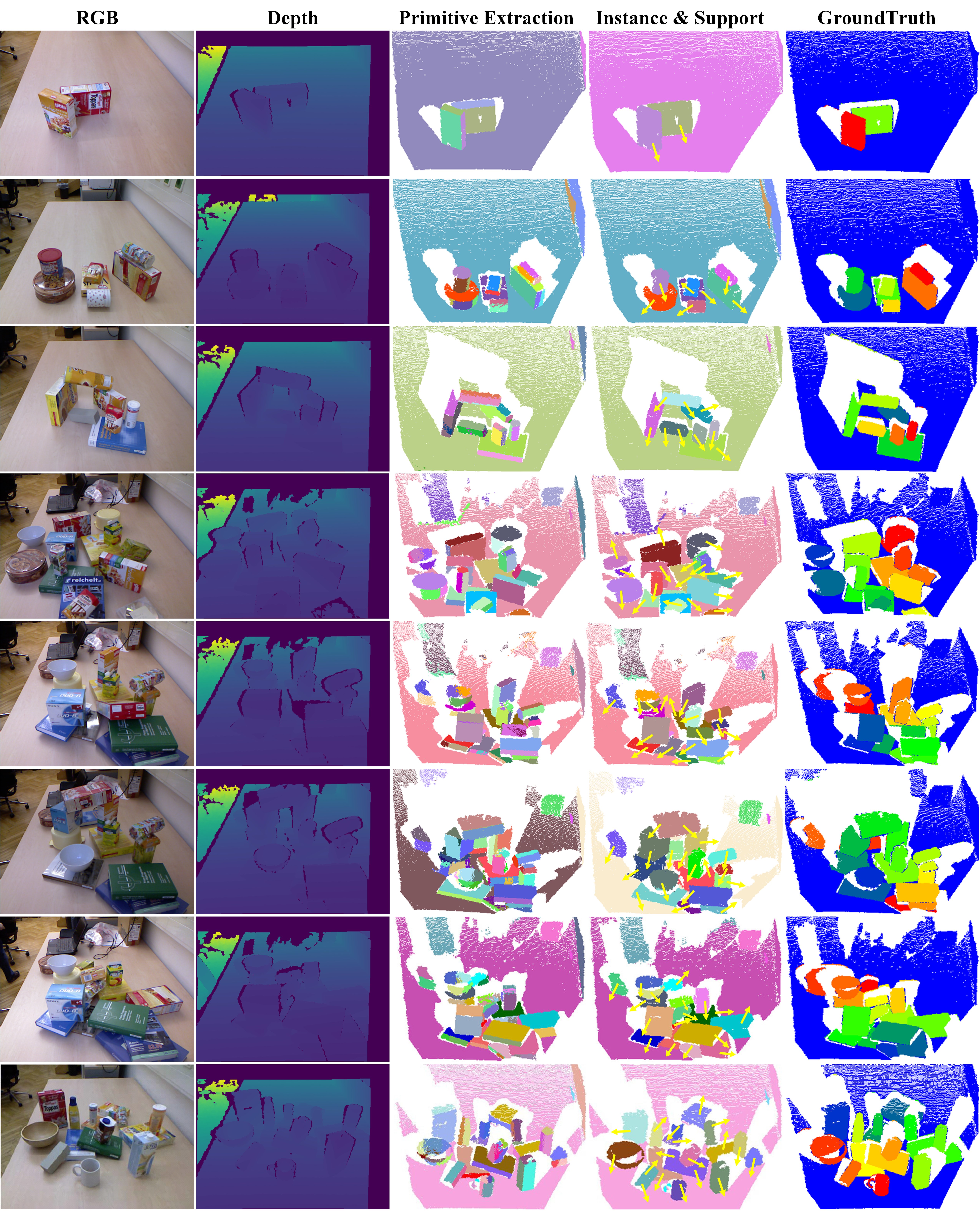}
\caption{\label{fig:segandsupport}
Experimental results of our algorithm. The first and second columns are RGB and depth images, respectively, from the OSD dataset. The middle column is the initial over segmentation of abstracting the scene to planar primitives. The primitive classification and support relations inference results are presented in the fourth column. The ground truth of the segmentation is shown in the last column.}
\end{figure}

Figure~\ref{fig:hgc} is an example of scene graph construction. In Figure~\ref{fig:hgc}, our method cannot detect the support relations between objects $C_{5}$ and $C_{15}$ due to occlusion, which is a key factor of errors in support relations inference. Some algorithms assume that an object can only be supported by one object when reasoning support relations \cite{yang2017support}\cite{zhuo2017indoor}. However, our method also considers that one object may be supported by two or more objects. For example, object $C_{12}$ is supported by both $C_1$ and $C_7$. Furthermore, our scene graphs are hierarchical. The primitive level contains pairwise spatial configurations of primitives in the adjacency graph, and the spatial configurations are divided into local support connection and local inner connection. The object level contains support relations between objects in the scene. The introduced root node ensures that any node can be traversed from the root node in the hierarchy graph. Figure~\ref{fig:segandsupport} shows more experimental results of support relations inference. 

\begin{table*}[htb]
\caption{Matrix of the scene graph. The red numbers indicate the errors from the scene graph in Figure~\ref{fig:hgc}. }
\label{tab:graph-matrix}
\begin{tabular}{llccccccccccccccccc}
\multicolumn{2}{l}{\multirow{2}{*}{}} & \multicolumn{17}{c}{\textbf{Supporting}} \\
\multicolumn{2}{l}{} & \multicolumn{1}{l}{$root$} & \multicolumn{1}{l}{$O_1$} & \multicolumn{1}{l}{$O_2$} & \multicolumn{1}{l}{$O_3$} & \multicolumn{1}{l}{$O_4$} & \multicolumn{1}{l}{$O_5$} & \multicolumn{1}{l}{$O_6$} & \multicolumn{1}{l}{$O_7$} & \multicolumn{1}{l}{$O_8$} & \multicolumn{1}{l}{$O_9$} & \multicolumn{1}{l}{$O_{10}$} & \multicolumn{1}{l}{$O_{11}$} & \multicolumn{1}{l}{$O_{12}$} & \multicolumn{1}{l}{$O_{13}$} & \multicolumn{1}{l}{$O_{14}$} & \multicolumn{1}{l}{$O_{15}$} & \multicolumn{1}{l}{$O_{16}$} \\ \cline{3-19} 
\multirow{17}{*}{\rotatebox{90}{\textbf{Supported}}} & \multicolumn{1}{l|}{$root$} & 0 & 0 & 0 & 0 & 0 & 0 & 0  & 0 & 0  & 0 & 0  & 0 & 0 & 0 & 0 & 0 & \multicolumn{1}{c|}{0} \\
 & \multicolumn{1}{l|}{$O_1$}  & 1 & 0 & 0 & 0 & 0 & 0 & 0 & 0 & 0 & 0  & 0  & 0  & 0 & 0  & 0 & 0 & \multicolumn{1}{c|}{0} \\
& \multicolumn{1}{l|}{$O_2$}  & 0 & 1 & 0 & 0 & 0 & 0 & 0 & 0 & 0 & 0  & 0  & 0  & 0 & 0 & 0  & 0 & \multicolumn{1}{c|}{0} \\
& \multicolumn{1}{l|}{$O_3$}  & 0 & 1 & 0 & 0 & 0 & 0 & 0 & 0 & 0  & 0 & 0  & 0 & 0 & 0 & 0  & 0  & \multicolumn{1}{c|}{0}\\
& \multicolumn{1}{l|}{$O_4$}  & 0 & 1 & 0 & 0 & 0 & 0 & 0 & 0 & 0 & 0 & 0 & 0 & 0 & 0  & 0  & 0 & \multicolumn{1}{c|}{0}\\
& \multicolumn{1}{l|}{$O_5$}  & 0 & 1 & 0 & 0 & 0 & 0 & 0 & 0 & 0  & 0  & 0  & 0 & 0 & 0  & 0  & 0 & \multicolumn{1}{c|}{0}\\
& \multicolumn{1}{l|}{$O_6$}  & 0 & 1 & 0 & 0 & 0 & 0 & 0 & 0 & 0 & 0  & 0 & 0 & 0 & 0  & 0 & 0  & \multicolumn{1}{c|}{0} \\
& \multicolumn{1}{l|}{$O_7$}  & 0 & 1 & 0 & 0 & 0 & 0 & 0 & 0 & 0 & 0  & 0  & 0  & 0  & 0 & 0 & 0 & \multicolumn{1}{c|}{0}\\
& \multicolumn{1}{l|}{$O_8$}  & 0 & 1 & 0 & 0 & 0 & 0 & 0 & 0 & 0 & 0 & 0 & 0 & 0 & 0 & 0 & 0 & \multicolumn{1}{c|}{0} \\
& \multicolumn{1}{l|}{$O_9$}  & 0 & 1 & 0 & 0 & 0 & 0 & 0 & 0 & 0 & 0 & 0 & 0 & 0 & 0 & 0 & 0 & \multicolumn{1}{c|}{0}     \\
& \multicolumn{1}{l|}{$O_{10}$} & 1/\textcolor{red}{0} & 0/\textcolor{red}{1} & 0 & 0 & 0 & 0 & 0 & 0 & 0 & 0 & 0 & 0 & 0 & 0 & 0 & 0 & \multicolumn{1}{c|}{0}     \\
& \multicolumn{1}{l|}{$O_{11}$} & 0  & 0 & 0  & 0 & 0 & 0 & 0 & 0 & 0 & 1 & 0 & 0 & 0 & 0 & 0 & 0 & \multicolumn{1}{c|}{0}     \\
& \multicolumn{1}{l|}{$O_{12}$} & 0 & 1 & 0 & 0 & 0 & 0 & 0 & 1 & 0 & 0 & 0 & 0 & 0 & 0 & 0 & 0  & \multicolumn{1}{c|}{0}    \\
& \multicolumn{1}{l|}{$O_{13}$} & 0 & 1 & 0 & 0 & 0 & 0 & 0 & 1 & 0 & 0 & 0 & 0 & 0  & 0 & 0 & 0 & \multicolumn{1}{c|}{0}     \\
& \multicolumn{1}{l|}{$O_{14}$} & 0 & 0 & 0  & 1 & 0 & 0 & 0 & 0 & 0 & 0 & 0 & 0 & 0 & 0 & 0 & 0 & \multicolumn{1}{c|}{0}     \\
& \multicolumn{1}{l|}{$O_{15}$} & 0 & 1 & 0 & 0 & 0 & 0/\textcolor{red}{1} & 0 & 0 & 0 & 0 & 0 & 0 & 0 & 0  & 0 & 0 & \multicolumn{1}{c|}{0}     \\
& \multicolumn{1}{l|}{$O_{16}$} & 0 & 0 & 0 & 0 & 0& 0& 0   & 0  & 0 & 0 & 0 & 0  & 1 & 0  & 0  & 0  & \multicolumn{1}{c|}{0}  \\ \cline{3-19} 
\end{tabular}
\end{table*}

We represent the directed and unweighted graph by its affinity matrix, as shown in Figure~\ref{fig:hgc} and Table~\ref{tab:graph-matrix}. Different from yang et al.\cite{yang2017support}, two significance rules are introduced here:
\begin{itemize} 
\item[1.] Supporting and supported relations of mis-segmented objects are ignored to magnify the impact of mis-segmentation.

\item[2.] Isolated nodes are supported by the root node, which ensures that the graph has no isolated node and the normalized Laplacian matrix is invertible\cite{chung1997spectral}.(defined in Eq.~\ref{eq-laplacian}).
\end{itemize}

Follow \cite{yang2017support},  Cheeger section and Spectral section are used to further evaluate the structure quality of our generated scene graphs(object level of scene hierarchy graphs). Essentially, they measure the similarity between generated scene graphs and ground truth graphs. The graph generated by our method is denoted as $G_{gs}$ while the ground truth is $G_{gt}$. Symmetric graph $G^{'}=(V^{'}, E^{'})$ is created from $G=(V, E)$ with undirected edges, as $V^{'}=V$ and $E^{'}=\{e^{'}_{ij} = 1 | e_{ij} = 1\  or\  e_{ji} = 1 \}$. $G_{gs}^{'}$ and $G_{gt}^{'}$ are created for the generated scene graph $G_{gs}$ and the ground truth $G_{gt}$.

For the Cheeger section and the Spectral section, the normalized Laplacian matrix is great important, which is defined as follows:
\begin{equation}
\label{eq-laplacian}
    \mathcal{L}_G = D_G^{-\frac{1}{2}}(D_G - A_G)D_G^{-\frac{1}{2}}
\end{equation}
where, $D_G$ and $A_G$ are the degree matrix and adjacency matrix of the graph $G$, respectively. Let $0=\lambda_0 < \lambda_1 \leq \cdots \leq \lambda_{n-1}$ be the eigenvalues of $\mathcal{L_G}$, and let $u_1$ be the eigenvector associated with eigenvalue $\lambda_1$ of $\mathcal{L}_G$.

The Cheeger constant of the graph, denoted $h_G$, is given by the following:

\begin{equation}
\label{eq:cheeger}
h_G(S) = \frac{|E(S, \bar{S})|}{min(volS, vol\bar{S})}
\end{equation}
where $S$ denotes a subset of the vertices of $G$, $volS = \sum_{x \in S} d_x$ being the volume of $S$, $d_x$ is the degree of vertex $x$, and $\bar{S} = V \setminus S$. The $|\cdot|$ denotes the cardinality. According to the inequality of the Cheeger constant in \cite{chung1997spectral}:

\begin{equation}
\label{eq:cheeger-inequality}
    \frac{1}{2}h_G^2 \le 1-\sqrt{1 - h_G^2} < \lambda_1 \le 2h_G
\end{equation}
the upper and lower bounds of the Cheeger constant can be inferred as follows:

\begin{equation}
\label{eq:cheeger-lu}
    l_G = \frac{1}{2} \lambda_1 \leq h_G <  \sqrt{2 \lambda_1} = u_G
\end{equation}
Here, $|(u_{G_{gs}^{'}} - l_{G_{gs}^{'}}) - (u_{G_{gt}^{'}} - l_{G_{gt}^{'}})|$
is used to evaluate the similarity between $G_{gs}^{'}$ and $G_{gt}^{'}$, called Cheeger section.

Another measure of the similarities between $G_{gs}^{'}$ and $G_{gt}^{'}$, called Spectral section, is denoted as follows:
\begin{equation}
\label{eq:spectral}
\parallel u_1(G_{gs}^{'}) \cdot u_1(G_{gs}^{'})^\top - u_1(G_{gt}^{'} \cdot u_1(G_{gt}^{'})^\top  \parallel_F / \sqrt{\mid V(G_{gt}^{'}) \mid}
\end{equation}
where $u_1$ is the eigenvector associated with the eigenvalue $\lambda_1$, $\parallel\cdot\parallel_F$ means the Frobenius-norm, $|\cdot|$ denotes the cardinality, and $\sqrt{|V(G_{gt}^{'})|}$ is for normalization.

Using Cheeger section and Spectral section, the evaluation results on the test dataset are shown in Table~\ref{gshg}. The generated scene graphs achieve low mean error values when comparing with ground truth. It proves that our bottom-up method is excellent at inferring support relations in 3D scenes. 

\begin{table}[htb]
\caption{Result of different measures to evaluate the quality of generated scene hierarchy graph comparing with ground truth.}
\label{gshg}
\centering
\begin{tabular}{c c c}
\toprule
Measures &Cheeger section(Eq.~\ref{eq:cheeger-lu}) & Spectral section(Eq.~\ref{eq:spectral})\\
\hline
Mean & 0.019 & 0.039 \\
Standard Deviation & 0.013 & 0.022\\
\bottomrule
\end{tabular}
\end{table}



\section{Conclusion}
\label{sec:conclusion}
In this study, we proposed a new approach to scene understanding, which utilized the topological information of primitive pairs in the scene, inferred the support relations in the scene bottom-up, and constructed a scene hierarchy graph. We demonstrated our approach's feasibility in experiments and excellent segmentation performance on an OSD dataset. Our approach could complement RGB-based approaches to scene understanding. In the future, we will attempt to use graph neural networks to learn more complex topological information of primitive pairs in 3D scenes. In addition, we will investigate how to generalize primitives from a plane to more base shapes. Finally, we will test whether the scene hierarchy graph we have constructed  would enable a robot to perform specific tasks.

\bibliographystyle{IEEEtran}
\bibliography{references.bib}


 




\vfill

\end{document}